\documentclass[pdflatex,sn-mathphys-num]{sn-jnl}

\usepackage{graphicx}
\usepackage{multirow}
\usepackage{amsmath,amssymb,amsfonts}
\usepackage{amsthm}
\usepackage{mathrsfs}
\usepackage[title]{appendix}
\usepackage{xcolor}
\usepackage{textcomp}
\usepackage{manyfoot}
\usepackage{booktabs}
\usepackage{algorithm}
\usepackage{algorithmicx}
\usepackage{algpseudocode}
\usepackage{listings}

\usepackage{eurosym}
\usepackage{multirow}
\usepackage{makecell}
\usepackage{float}
\usepackage{caption}
\usepackage{subcaption}
\usepackage{multicol}
\usepackage{hyperref}
\usepackage{longtable}

\usepackage[absolute,overlay]{textpos}
\usepackage{boxedminipage}
\usepackage[export]{adjustbox}% http://ctan.org/pkg/adjustbox

\begin{document}

\title{Global urban visual perception varies across demographics and personalities}
\author[1]{\fnm{Matias} \sur{Quintana}}\email{matias.quintana@sec.ethz.ch}
\author[1,2]{\fnm{Youlong} \sur{Gu}}\email{youlong@u.nus.edu}
\author[2]{\fnm{Xiucheng} \sur{Liang}}\email{xiucheng@u.nus.edu}
\author[2]{\fnm{Yujun} \sur{Hou}}\email{yujun@nus.edu.sg}
\author[2]{\fnm{Koichi} \sur{Ito}}\email{koichi.ito@u.nus.edu}
\author[2]{\fnm{Yihan} \sur{Zhu}}\email{yihan.zhu@u.nus.edu}
\author[2]{\fnm{Mahmoud} \sur{Abdelrahman}}\email{e0348764@u.nus.edu}
\author*[2,3]{\fnm{Filip} \sur{Biljecki}}\email{filip@nus.edu.sg}

\affil[1]{\orgdiv{Future Cities Lab Global}, \orgname{Singapore-ETH Centre}, \orgaddress{\street{CREATE campus, 1 Create Way, \#06-01 CREATE Tower}, \city{Singapore}, \postcode{138602}, \country{Singapore}}}

\affil[2]{\orgdiv{Department of Architecture}, \orgname{National University of Singapore}, \orgaddress{\street{4 Architecture Dr}, \city{Singapore}, \postcode{117566}, \country{Singapore}}}

\affil[3]{\orgdiv{Department of Real Estate}, \orgname{National University of Singapore}, \orgaddress{\street{15 Kent Ridge Drive}, \city{Singapore}, \postcode{119245}, \country{Singapore}}}

\abstract{ %8-10 sentences
\begin{textblock*}{\textwidth}(3.8cm,1.1cm) 
\begin{center}
\begin{footnotesize}
% \vspace*{-13cm}
\begin{boxedminipage}{1\textwidth}
The version of record of this article, first published in \emph{Nature Cities}, is available online at Publisher’s website: \url{http://dx.doi.org/10.1038/s44284-025-00330-x}\\ Cite as:
Quintana M, Gu Y, Liang X, Hou Y, Ito K, Zhu Y, Abdelrahman M, Biljecki F (2025): Global urban visual perception varies across demographics and personalities. \textit{Nature Cities}.
\end{boxedminipage}
\end{footnotesize}
\end{center}
\end{textblock*}
Understanding people's preferences is crucial for urban planning, yet current approaches often combine responses from multi-cultural populations, obscuring demographic differences and risking amplifying biases. 
We conducted a largescale urban visual perception survey of streetscapes worldwide using street view imagery, examining how demographics -- including gender, age, income, education, race and ethnicity, and personality traits -- shape perceptions among 1,000 participants with balanced demographics from five countries and 45 nationalities. 
This dataset, Street Perception Evaluation Considering Socioeconomics (SPECS), reveals demographic- and personality-based differences across six traditional indicators -- safe, lively, wealthy, beautiful, boring, depressing -- and four new ones -- live nearby, walk, cycle, green. 
Location-based sentiments further shape these preferences. 
Machine learning models trained on existing global datasets tend to overestimate positive indicators and underestimate negative ones compared to human responses, underscoring the need for local context. 
Our study aspires to rectify the myopic treatment of street perception, which rarely considers demographics or personality traits.
}


\maketitle

%%==================================%%
\section{Introduction}\label{sec:intro}

The urban environment fundamentally shapes human perception and behavior in cities~\cite{Dijksterhuis.2001}, with visual elements influencing both immediate perception and deeper cognitive responses to spaces~\cite{Yang2024}.
Urban visual perception -- how people interpret and respond to visual information in city environments -- has become central to understanding urban experiences~\cite{carmona2021public, gibson_ecological_2014, Zhang.2018j3c}, particularly as Street View Imagery (SVI) has revolutionized this research domain~\cite{Dubey.2016, ito_understanding_2024}.
Large-scale online surveys using SVI have largely replaced traditional on-site fieldwork~\cite{ito_understanding_2024}, significantly expanding geographic scope and participant diversity~\cite{Dubey.2016, Lefosse.2025}.

Building on this methodological evolution, the MIT Place Pulse projects~\cite{Salesses.2013, Dubey.2016} stand out for their unprecedented scale. 
Place Pulse 2.0 (PP2) collected ratings for over 110,000 street view images from 56 cities across six continents, evaluated by participants from 162 countries using pairwise comparisons for six perceptual indicators: \textit{safe}, \textit{lively}, \textit{wealthy}, \textit{beautiful}, \textit{boring}, and \textit{depressing}.
While machine learning models trained on these comparisons have advanced perception studies in multiple urban regions~\cite{Zhang.2018j3c, ji_new_2021, Kang.2023, Rui.2025}, these multi-city models may introduce bias or mask nuances in subjective preferences.

These demographic and location-specific differences are crucial for effective urban planning. 
Gender affects the perception of safety~\cite{hidayati_how_2020, Cui.2023c2p, Dubey.2025, Zhou.2025krn}, with women generally perceiving lower safety in scenes with overall low safety scores~\cite{Cui.2023c2p} and focusing on areas outside the walking path~\cite{Chaney.2024ujk}, and responding differently to green spaces~\cite{Jiang.2014xp}. 
These differences may be important for informing inclusive design and strategies to mitigate urban heat challenges~\cite{Kousis.2022, Bartesaghi-Koc.2020, Ananyeva.2023, Rafiee.2016}.
These studies revel helpful group-based preferences for context-aware solutions~\cite{Qi.2022} and equality initiatives~\cite{Carpio-Pinedo.2019nyb}.

Despite extensive research on demographic factors in urban planning~\cite{Jiang.2014xp, Carpio-Pinedo.2019nyb, kazemi_peoples_2023, Kang.2023}, most studies remain limited to single cities or specific demographics, with narrow perceptual focus. 
Recent work has expanded geographic scope, e.g., examining biophilic perception across eight cities~\cite{Lefosse.2025} or comparing resident surveys with multi-city model predictions in Stockholm~\cite{Kang.2023}, but typically addresses single perceptual dimensions with minimal demographic granularity.
Additionally, personality factors, which influence environment interpretation~\cite{yang2024influences}, urban spaces perception~\cite{yang2024influences}, and quality of life~\cite{pocnet2017personality, huang2017does}, remain under-explored in visual urban perception research.
Urban science thus lacks a comprehensive study examining, simultaneously, multiple locations with diverse participants, including personality characteristics, across various perception dimensions. 
While perception is inherently personal, understanding how much is driven by demographics and personality remains a critical research gap.

Our study analyzes urban visual perception across five continents through 1,000 demographically diverse participants from Santiago, Chile; the greater Amsterdam area, Netherlands; Nigeria; Singapore; and San Francisco and Santa Clara, USA, who rated 400 street-level images from their own or neighboring cities, addressing critical gaps in previous research.
Unlike most urban perception studies, we prioritized acquiring comprehensive demographic characteristics~\cite{Zhu.2025}, including gender, age, education, income, race/ethnicity, and personality traits (extraversion, agreeableness, conscientiousness, neuroticism, and openness). 
Building on previous smaller-scale deployments~\cite{Quintana.2024}, we expanded both geographic scope and perceptual dimensions by introducing four new indicators (\textit{live nearby}, \textit{walk}, \textit{cycle}, and \textit{green}) to complement the traditional six (\textit{safe}, \textit{lively}, \textit{wealthy}, \textit{beautiful}, \textit{boring}, and \textit{depressing}).

Our statistical analyses identify how demographic characteristics and personality traits moderate perception across multiple cities, while also examining magnitude bias propagation in widely-used machine learning models trained on the PP2 dataset, which are frequently applied to specific urban contexts such as Shanghai and Beijing~\cite{Zhang.2018j3c}, Boston and Los Angeles~\cite{Kang.2021guq}, Amsterdam~\cite{gao_pedaling_2025}, and New York and Singapore~\cite{wang_assess_2024}.
We conducted comparative analyses in two key scenarios: multi-city imagery rated by single-city participants and single-city imagery rated by multi-city participants. 
Additionally, we quantified how participants rated their own cities compared to others, revealing critical differences in perception when using non-local imagery for localized evaluation.
Lastly, we validated our new proposed indicators by analyzing their distinctiveness and complementary relationships with the existing six, highlighting their potential to support ongoing urban research questions on liveability, cyclability, walkability, and greenery.

\begin{figure}[tbp]
    \centering
    \includegraphics[width=1\textwidth]{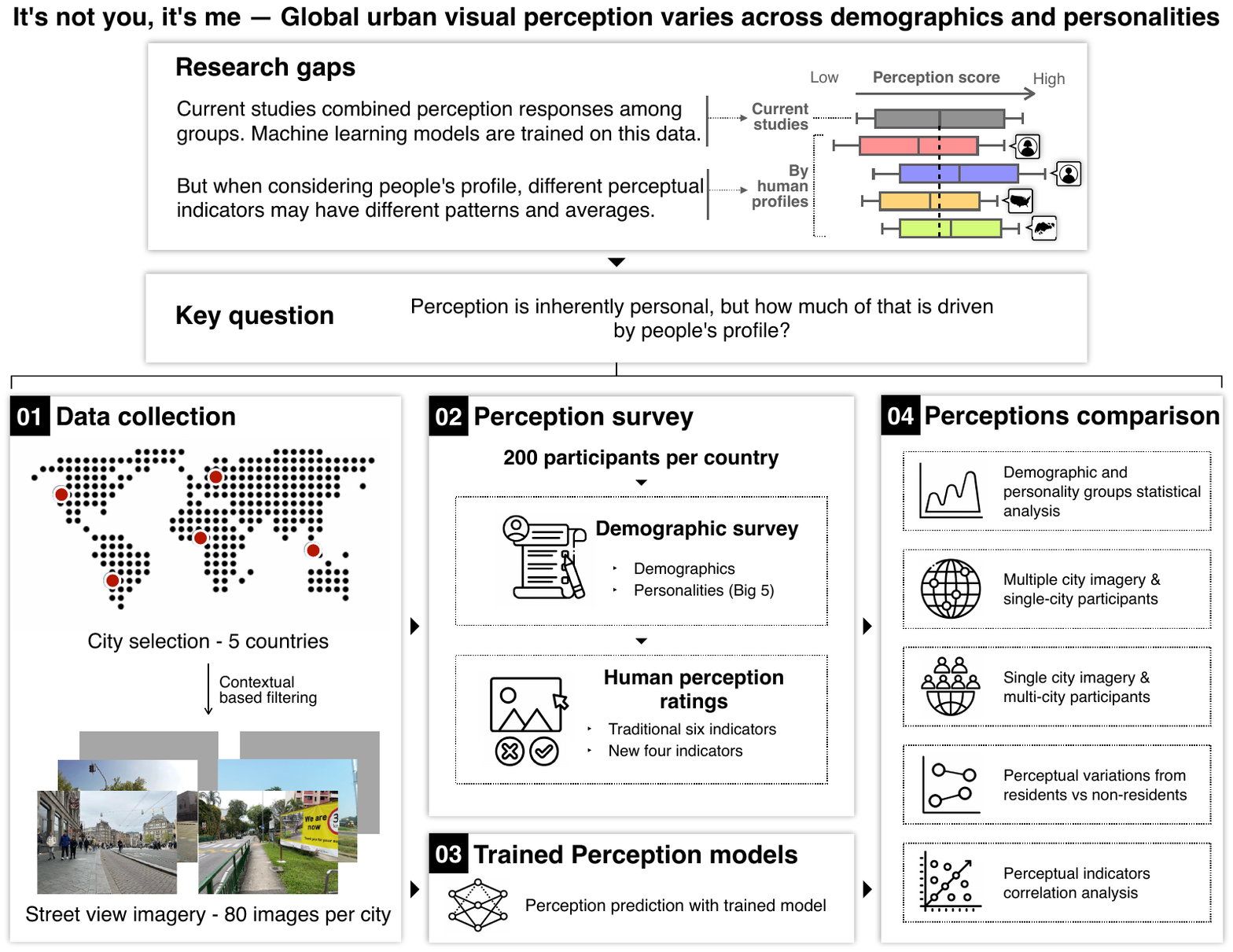}
    \caption{
    \textbf{Research gaps and key questions in urban visual perception studies, followed by our methodology to bridge them.}
    The profile survey included demographics (gender, age group, education level, annual household income, and race and ethnicity) and personality traits (extraversion, agreeableness, conscientiousness, neuroticism, openness).
    The human perception ratings included the traditional six indicators (\textit{safe}, \textit{wealthy}, \textit{lively}, \textit{beautiful}, \textit{depressing}, and \textit{boring}) and new proposed four indicators (\textit{live nearby}, \textit{walk}, \textit{cycle}, and \textit{green}) for pairwise comparisons.
    Source of imagery: Mapillary and KartaView contributors.
    We acknowledge the use of icons from The Noun Project, created by various authors and licensed under CC BY 3.0.
    }
    \label{fig:methodology-panel}
\end{figure}

Fig.~\ref{fig:methodology-panel} provides an overview of the research gaps, the key research question, and our methodology. 
Through this comprehensive set of analyses and by making our dataset and codebase publicly available, this work contributes substantially to diversity and representation in SVI perception studies while establishing a foundation for context-specific, human-centric urban visual perception research.

%%===================================================%%
\section{Results}
\subsection{Demographics and personalities as moderators of perception}
\begin{figure}
    \centering
    \includegraphics[width=1\linewidth]{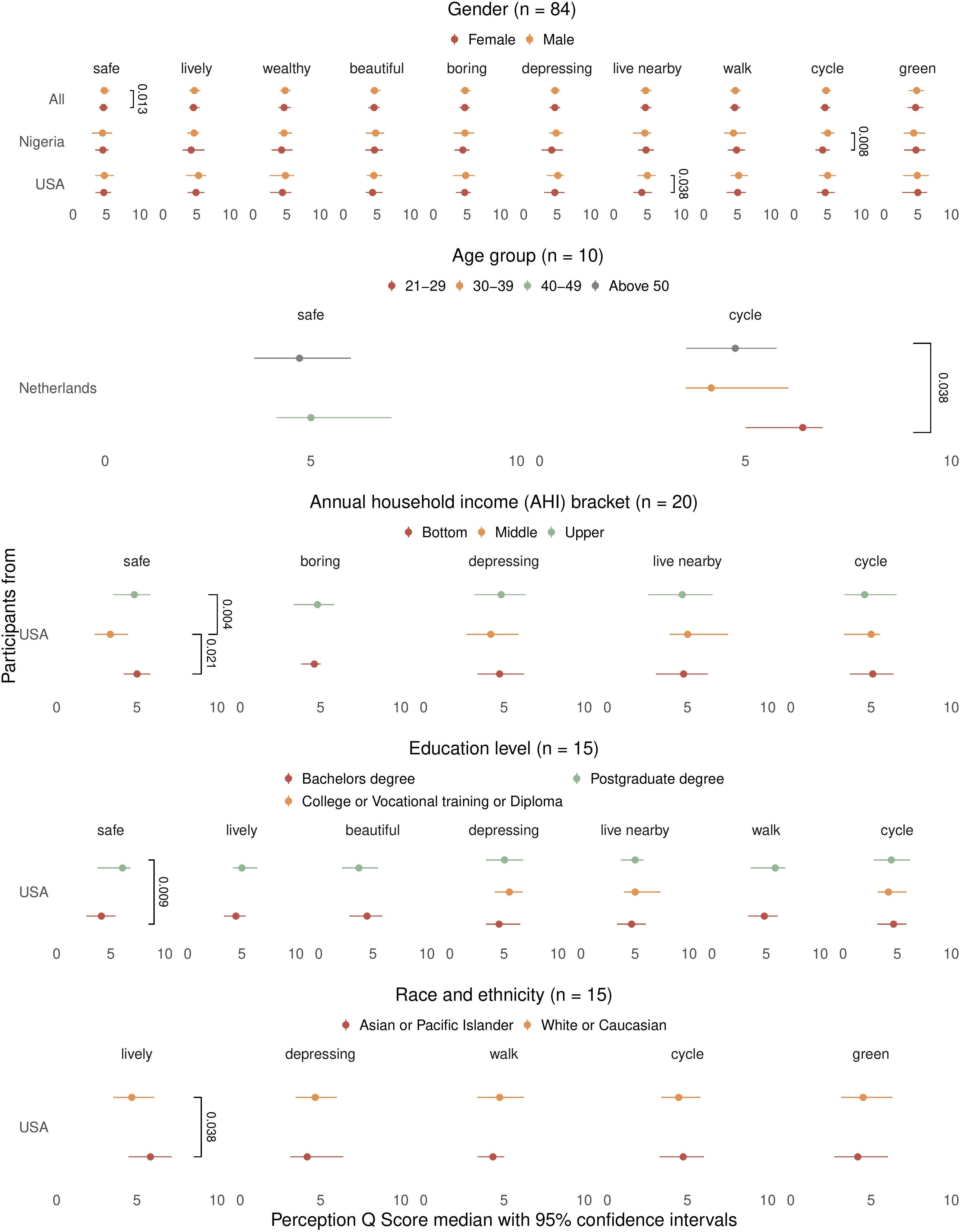}
    \caption{
    \textbf{Statistical difference of perception scores by demographics.}
    Perception Q scores are calculated from ratings by participants in each group for all (without location grouping) and each location.
    Welch's ANOVA was used for demographic attributes with only two groups and for demographic attributes with more than two groups, we performed the Games-Howell post-hoc test.
    Minimum sample size $n$ (rated images with at least four pairwise comparison by participants per group) is shown for each demographic profile.
    Locations with no significant differences in any indicator and demographic groups with fewer than $n$ samples are not shown.
    Significance thresholds at $p<0.05$.
    }
    \label{fig:demographic-stats}
\end{figure}

We evaluated whether the mean SVI perception Q scores significantly differ across participants grouped by demographic characteristics and personality traits.
We found differences between demographic and personalities groups but they differ by locations, i.e., where participants are from, and perceptual indicators (\autoref{fig:demographic-stats}).
The highest number of significant differences happened in populations grouped by gender, and across all five demographics (gender, age group, annual household income bracket, education level, and race and ethnicity) the traditional indicator \textit{safe} had the most intra-country differences, followed by our new indicator \textit{cycle}.
No significant differences were found in negative indicators and only one significant difference was found in the \textit{live nearby} and \textit{lively} indicator in gender and race and ethnicity groups, respectively, in the USA.
In terms of countries, we found differences in participants from San Francisco and Santa Clara (USA) in four out of the five demographic characteristics and no differences in any demographic group of participants from Santiago (Chile) or Singapore.
We noted gender differences under the \textit{live nearby}, and \textit{cycle} indicators for specific countries (USA and Nigeria, respectively), and across all locations (e.g., Q scores calculated without grouping by countries) under the \textit{safe} indicator (top plot in \autoref{fig:demographic-stats}).

Differences between age groups occurred mostly among participants in the ``21-29'' and ``Above 50'' groups (second plot from the top \autoref{fig:demographic-stats}).
This group had differences in perception for the \textit{cycle} indicator in Netherlands.
In groups formed by Annual Household Income (AHI) brackets, we noticed the difference between participants in the ``Bottom'' bracket in relation to the ``Middle'' and ``Upper'' brackets for perception scores in the \textit{safe} indicator in the USA (middle plot \autoref{fig:demographic-stats}).

Differences among participants' education levels were only found in participants from the USA. 
These differences happened between bachelor's and postgraduate degree holders for the \textit{safe} indicator (fourth plot from the top ~\autoref{fig:demographic-stats}).
Lastly, we found differences in groups determined by the participants' self-reported race and ethnicity in the USA (bottom plot in \autoref{fig:demographic-stats}).
The differences for the \textit{lively} indicator appeared between ``Asian or Pacific Islanders'' and ``White or Caucasian'' participants.
The demographic groups on our dataset are consistent with the population mix in the sampled cities of San Francisco and Santa Clara (USA)

While overall we found fewer differences within demographic interaction nested groups, i.e., a total of four compared to eight in single demographic groups (\autoref{fig:demographic-stats}), among all the combinations of `gender $\times$ age group $\times$ AHI' nested groups, half the differences happened within-country and half across all locations (Extended data Fig. 1).
Differences formed by all three demographics combined were found for the \textit{lively} indicator between ``Female x 30-39 x Bottom'' and ``Male x 21-29 x Middle'' nested groups only when all participants' responses were combined, although under a small sample size ($n \geq 5$, top plot in Extended Data Fig. 1).
Within `gender $\times$ age group' nested groups perception differences for the \textit{walk} indicator between genders of the ``Above 50'' age group in the USA also under a smaller sample size ($n \geq 5$, second plot from the top Extended data Fig. 1).
Then, differences for the \textit{safe} indicator between genders of the ``Middle'' AHI group were found in Netherlands under a larger sample size.
Finally, perceptual differences between age groups ``21-29'' and ``30-39'' of the ``Bottom'' AHI bracket were found for the \textit{lively} indicator across all locations.
Both perceptual differences for \textit{safe} and \textit{lively} were found with higher sample size ($n \geq 20$, last two bottom plots Extended data Fig. 1, respectively).

\begin{figure}
    \centering
    \includegraphics[width=1\textwidth]{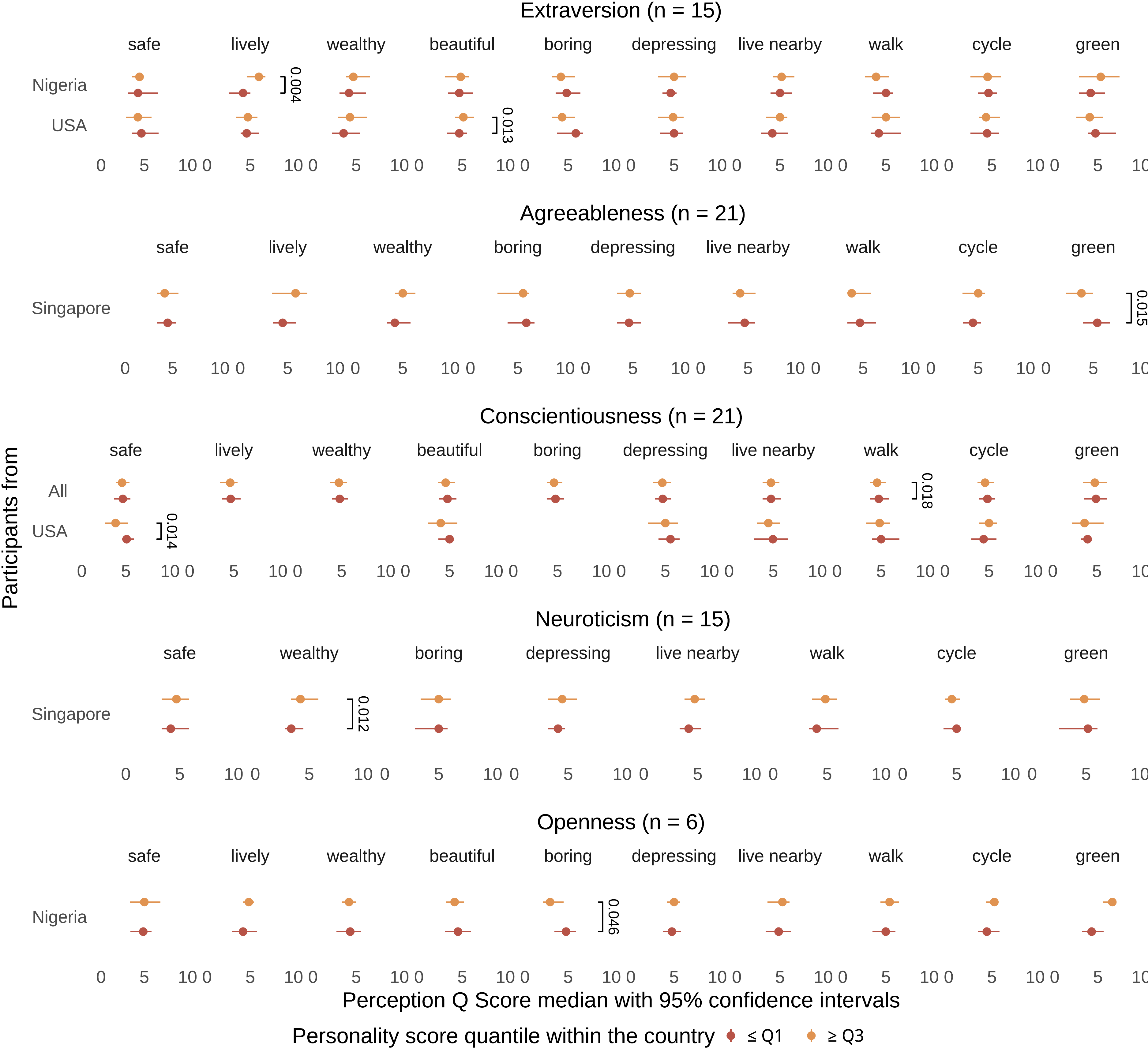}
    \caption{
   \textbf{Statistical difference of perception scores by personality traits.}
    Perception Q scores are calculated from ratings by participants in each group for all (without location grouping) and each location.
    We performed the Games-Howell post-hoc test and the minimum sample size $n$ (rated images with at least four pairwise comparison by participants per group) is shown for each personality trait $\leq$Q1 and $\geq$Q3 quartile.
    Locations with no significant differences in any indicator and personality groups with fewer than $n$ samples are not shown.
    Significance thresholds at $p<0.05$.
    }
    \label{fig:personalities-stats}
\end{figure}

We also found differences in participants with the most distinct personality traits, i.e., when their intra-country personality trait score was $\leq$Q1 or $\geq$Q3, i.e., within the 25$^{th}$ and 75$^{th}$ percentiles, respectively.
The most significant differences happened in populations grouped by their extraversion and conscientiousness score, with similar number of overall differences compared to the single demographic analysis (seven and eight differences, respectively).
Unlike the single demographic analysis, one significant difference was found under a negative indicator, \textit{boring}, between personality scores in openness in Nigeria, albeit with the lowest sample size among the personality groups ($n \geq 5$, bottom plot \autoref{fig:personalities-stats}).
Similarly to the demographic analysis, we did not found differences in participants from Santiago (Chile) but found two difference in two personality groups in participants from Singapore.

Perceptual differences among participants with high and low personality scores in extraversion exist for the \textit{lively} and \textit{beautiful} indicators in Nigeria and San Francisco and Santa Clara (USA) (top plot \autoref{fig:personalities-stats}).
Among participants with agreeableness traits, we found differences in the new \textit{green} indicator in Singapore (seconc plot from the top \autoref{fig:personalities-stats}).
Differences between participants with conscientiousness personality traits are notable for the \textit{safe} and \textit{walk} indicators in participants from San Francisco and Santa Clara (USA) and across all locations, respectively (middle plot \autoref{fig:personalities-stats}).
Lastly, we saw one difference in the personality trait of neuroticism for the \textit{wealthy} indicator in Singapore (fourth plot from the top \autoref{fig:personalities-stats}).

\subsection{Multi-city imagery \& single-city participants}

\begin{figure}
    \centering
    \includegraphics[width=\linewidth]{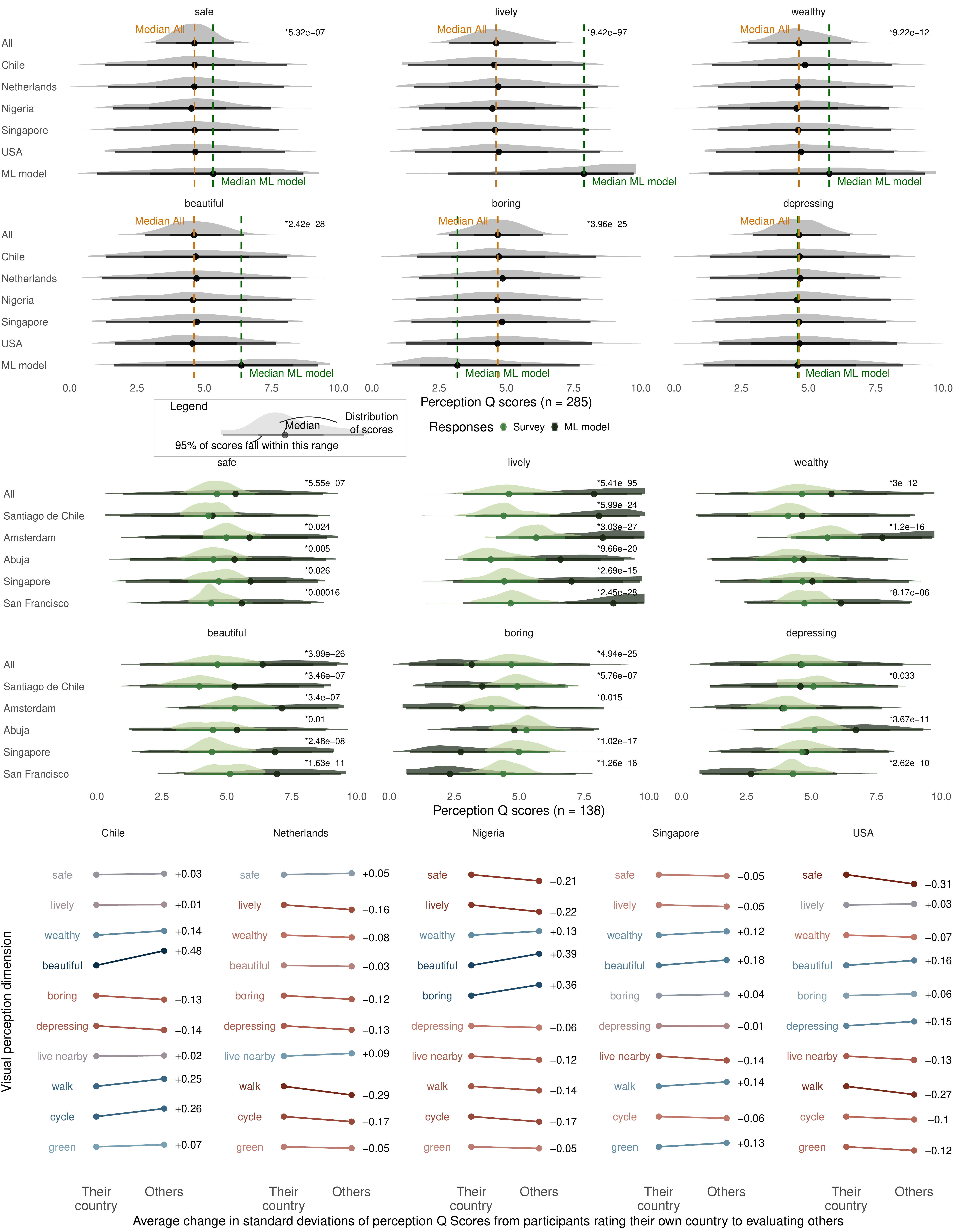}
    \caption{
    \textbf{Street-level images perception Q scores statistical comparisons based on where people are from (top), where SVIs were taken (middle), and rated by people from their own versus other cities (bottom)}.
    Welch's ANOVA was used to compared the combined scores by location (All) against the ML model predictions (top) and to compare scores by each location against the ML model predictions (middle).
    Minimum sample size $n$ (rated images with at least four (top) or 22 (middle) pairwise comparison by participants per location) is shown.
    Significance threshold at *$p<0.05$.
    }
    \label{fig:perception-comparisons}
\end{figure}

We looked at how people living in the same place perceive different cities and compared them with the perceptual predictions of an off-the-shelf deep learning model fine-tuned with a global dataset (ViT-PP2).
Overall, we found small but noticeable differences in perception scores given by participants in different countries, and that ViT-PP2 perceptual predictions over and underestimate positive and negative indicators, respectively (top plot in \autoref{fig:perception-comparisons}).

The distribution and spread of perception Q scores, calculated by where participants are from, shows little variability between countries and indicators.
We noted the differences when focusing on the median of each distribution and comparing it with the score median of all combined ratings (black dot and orange vertical dashed line in the top plot in \autoref{fig:perception-comparisons}).
In particular, we see lesser variability in the median scores for the \textit{safe} and \textit{depressing} indicators, as noted by the black filled circles being closely vertically aligned with the orange dashed line.

The perception scores given by the ViT-PP2 model do not follow this trend.
These scores are overestimated, with median scores 15\%, 71\%, 24\%, and 38\% higher than the combined (All) median, in positive indicators -- \textit{safe}, \textit{lively}, \textit{wealthy}, and \textit{beautiful}, respectively.
Conversely, the scores are underestimated with median scores as low as 32\% lower than the combined (All) median, in negative indicators -- \textit{boring} and \textit{depressing}.
The difference in the latter is much lesser, with the combined (All) median Q Score being 1\% lower than the ViT-PP2 median, and the only difference that is not statistically significant (bottom left subfigure on top plot in \autoref{fig:perception-comparisons}).

\subsection{Single-city imagery \& multi-city participants}

In our third analysis, we compared how people from all five different countries perceived the same city and also compared these perception scores with the perception predictions from ViT-PP2.
Unsurprisingly, we found that different cities are perceived differently, e.g., different median perception scores for all cities across indicators.
However, we also found that the difference between a city's SVI median perception score and ViT-PP2 predicted score fluctuates depending on the indicator.
Similarly to the previous analysis, we found that ViT-PP2 perceptual predictions over and underestimate positive and negative indicators, respectively (middle plot in \autoref{fig:perception-comparisons}).

The difference between perception Q scores distributions of the different cities' SVI is more visually apparent here, and we see the variability for their median score in each indicator by following the position of the green-filled circles, as they are far from being vertically aligned.
We found the highest difference, across all countries, between the ViT-PP2 median predicted perception and the city's SVI median perception, from 45\% to an 84\% increase, for the \textit{lively} indicator.
The smaller differences, from 9\% decrease to 3\% increase, are in the \textit{depressing} indicator, with exceptions for SVI from Abuja (31\% increase) and San Francisco (37\% decrease).
Not surprisingly, the smallest differences in Amsterdam and all combined countries are the only differences that are not statistically significant in this indicator.
The \textit{safe} and \textit{wealthy} indicators also show countries with small differences between their ground truth and the predicted scores.
For the \textit{safe} indicator, we saw a 3\% increase in Santiago, while for the \textit{wealthy} indicator, a 13\% increase alongside an approximately 8\% increase in Abuja and Singapore.
None of these differences were statistically significant.

As we encountered in the previous analysis, comparing perception scores of multi-city SVI from single-city participants, the perception scores predicted by the ViT-PP2 model overestimate positive indicators (\textit{safe}, \textit{lively}, \textit{wealthy}, and \textit{beautiful}), and underestimate negative ones (\textit{boring} and \textit{depressing}).
We observe this as the black-filled circle (model median) is on the right of the green-filled circle (country SVI median) for overestimates and vice-versa for underestimates, and most differences are statistically significant (middle plot in \autoref{fig:perception-comparisons}).
Similarly, we found some exceptions in the \textit{depressing} indicator where the median values are almost overlapping in Singapore, Amsterdam, and all combined cities.

\subsection{Perceptual variation from residents vs non-residents}

As the last comparison of perception Q scores, we expanded on the second analysis on multiple-city SVI and single-city participants and looked into how people living in the same, or neighboring, city perceive other cities relative to their own.
We analyzed the average change in the perception scores, scaled to z-scores within participants' locations to debias responses from cultural and location-based influences, of participants who rated the SVI of their own city compared to the SVI of the other four cities.
We found that some populations perceived their city mostly better (higher positive indicators scores) and others perceived their city overall worse (lower positive indicator and higher negative indicator scores) (bottom plot in \autoref{fig:perception-comparisons}).
In this figure, the differences shown on the slopes are based on z-scores and allow us to compare how different a response is when compared to all responses, i.e., standard deviation units (SDs).
This format allows us to avoid the participants' bias that could appear when using absolute perception Q scores.

More than 78\% participants in each country have lived in their respective cities -- Santiago (Chile), the greater Amsterdam area (Netherlands), Nigeria (multiple cities), Singapore, and San Francisco and Santa Clara (USA) -- for more than five years, and less than 5\% of them for less than one year (Supplementary material Table 1).
This time spent living in the city suggests that the participants would have assimilated into the local culture, customs, and behaviors, and their responses can be regarded as those of someone originally from that area.

Participants from the greater Amsterdam region have a mostly better perception of Amsterdam, compared to the other four cities.
These participants rated the other four cities, on average, between 0.03 and 0.29 SDs lower in positive indicators \textit{lively}, \textit{wealthy}, and \textit{beautiful}, and in our new proposed indicators \textit{walk}, \textit{cycle}, and \textit{green}.
The indicator \textit{safe} and \textit{live nearby} were the exception where the other four cities were perceived, on average, 0.05 and 0.09 SDs higher, respectively.
They also rated Amsterdam, on average, 0.12 and 0.13 SDs more \textit{boring} and \textit{depressing}, respectively.

Participants from San Francisco and Santa Clara also rated San Francisco SVI higher (0.07 to 0.31 SDs more) on average in all positive indicators and all new indicators, except for \textit{lively} and \textit{beautiful}.
They also rated other cities as more boring and depressing on average (0.06 and 0.15 SDs, respectively).
In Singapore, the results are more mixed.
The city is perceived on average slightly higher and lower in positive indicators.
The indicators \textit{safe} and \textit{lively} are perceived 0.05 SDs higher, and the indicators \textit{wealthy} and \textit{beautiful} are perceived 0.12 and 0.18 SDs lower.
The city is also perceived, on average, as less boring (0.04 SDs) and slightly more depressing (0.01 SDs).
Lastly, participants in Singapore also perceived their city, on average, as less walkable and green (0.14 and 0.13 SDs less), 0.06 SDs more cycleable, and 0.14 SDs more suitable to \textit{live nearby} than other cities.
Interestingly, these results are consistent with the city's geographic characteristics of tropical weather and high temperatures.
Despite the city's abundant greenery, this is not reflected in the \textit{green} scores.
Results in Nigeria follow a similar mixed trend.
Participants in Nigeria rated Abuja, on average, around 0.22 SDs more \textit{safe} and \textit{lively} and 0.13 and 0.39 SDs less \textit{wealthy} and \textit{beautiful}, respectively.
The city is also perceived 0.36 SDs less \textit{boring} and 0.06 SDs more \textit{depressing}.
Among the new indicators, the city was perceived, on average, between 0.05 and 0.17 SDs higher than \textit{live nearby}, \textit{walk}, \textit{cycle}, and \textit{green}.

On the other hand, results from participants in Santiago suggest its participants perceive them mostly worse: lower in positive indicators and higher in negative indicators.
The city was perceived as 0.01 to 0.48 SDs lower in all positive indicators, including our new four proposed indicators, and around 0.14 higher in all negative indicators.
The highest differences are in the \textit{beautiful} indicator, with scores for the other cities being 0.48 SDs higher on average.

\subsection{Perceptual indicators correlation}
\begin{figure}[t]
    \centering
    \includegraphics[width=1\linewidth]{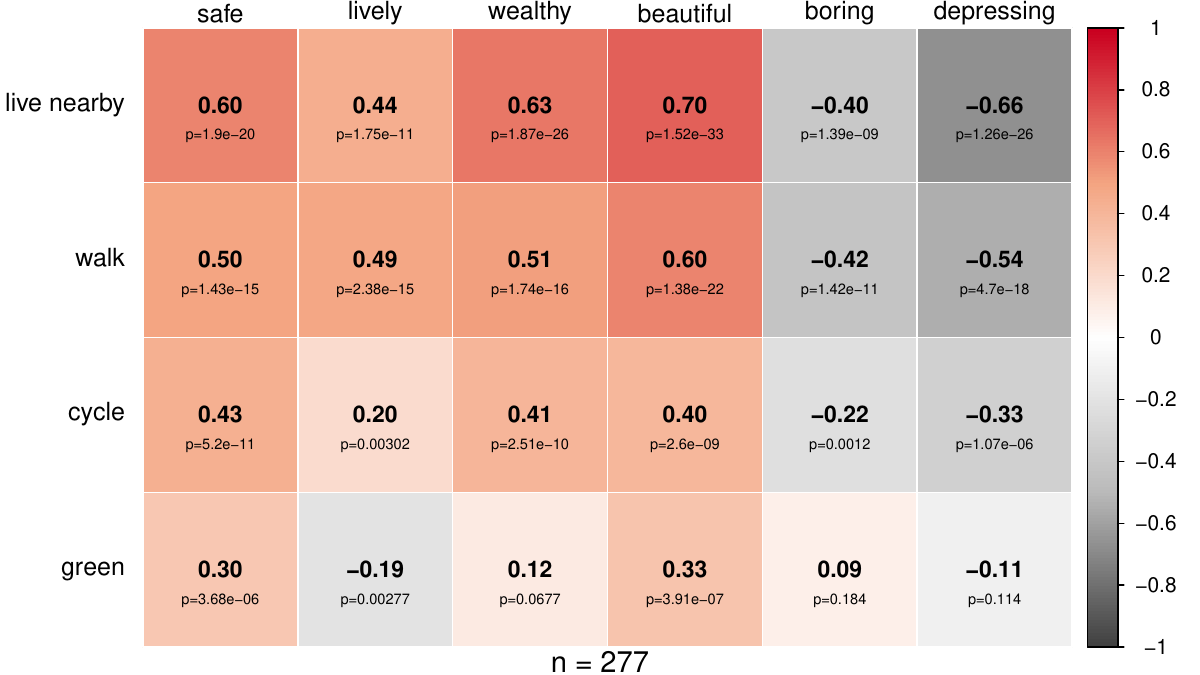}
    \caption{
    \textbf{Correlation between our four new indicators (rows) and the six predominantly used indicators (columns) across all locations.}
    Pearson correlation R-values where all correlations are significant ($p<0.05$) except for the correlations of the \textit{green} indicator and the positive indicator \textit{wealthy} and negative indicators \textit{boring} and \textit{depressing}.
    Minimum sample size $n$ (rated images with at least 22 pairwise comparison by participants per indicator) is shown.
    }
    \label{fig:corr}
\end{figure}

We performed a linear correlation analysis and found that our new proposed indicators correctly capture the positive and negative sentiment of the existing perception indicators.
The new indicators \textit{live nearby}, \textit{walk}, and \textit{cycle} show a positive correlation (R$>$0.40) with the traditional positive indicators (\textit{safe}, \textit{lively}, \textit{wealthy}, and \textit{beautiful}) and a negative correlation (R$<$-0.40) with the negative indicators (\textit{boring} and \textit{depressing}) (\autoref{fig:corr}).
An exception is seen in the relationship between \textit{cycle} and \textit{lively} (R$>$0.20) and \textit{cycle} and \textit{boring} (R$<$-0.20).
The newly proposed \textit{green} indicator shows an inverse result of negative correlation with a positive indicator (\textit{lively}) and vice versa (\textit{boring}).
An extended linear correlation analysis is shown in Extended data Fig. 2.

We further investigated non-linear relationships with polynomial regressions.
Extended data Fig. 3 shows the best fitting curve in red (linear, quadratic, or cubic) in terms of $R^2$, for the different perceptual pairs, i.e., new indicator vs traditional indicator.
In most instances, the best fitting model for \textit{live nearby}, \textit{walk}, and \textit{cycle} is a linear model (in three, five, and four out of the six perceptual pairs, respectively).
In the remaining perceptual pairs, the best fitting model has, in all but one case, an $R^2$ value 0.1 higher than the respective linear model, with the highest difference being 0.2 in the \textit{live nearby} vs \textit{wealthy} pair.
This suggests that these three new perceptual indicators are predominately linearly correlated with the traditional six indicators.
The \textit{green} perceptual indicator appears to have a more complex relationship with the traditional six indicators (Extended data Fig. 3d).
This indicator best correlates linearly with \textit{safe} and \textit{lively}, R values of 0.30 and -0.19, respectively and $R^2$ values of 0.09 and 0.04.
There is a weak non-linear relationship between the \textit{green} perceptual indicator and \textit{wealthy}, \textit{boring}, and \textit{depressing} indicators ($0.03\leq R^2\leq0.05$), though still three times higher than $R^2$ values from linear models.
Nevertheless, the relationship between \textit{green} and \textit{beautiful}, although relatively low ($R^2 = 0.12$ with a cubic model and $R^2$ = 0.11 with a linear model), is as much as three times higher in $R^2$ values than the other correlations.

%%===================================================%%
\section{Discussion}\label{sec:discussion}
Across all cities chosen, the statistical analysis revealed that differences in global urban visual perception are explained by demographics and personalities (\autoref{fig:demographic-stats}).
While existing work found safety perceptions, in specific locations, are explained by gender~\cite{Gu.2025, Zhu.20243e, Dubey.2025, Cui.2023c2p, Zhou.2025krn}, our findings revealed a potential aggregated difference when considering all five locations together.
Similar to work done in \cite{Kang.2023}, we did not find statistically significant differences between age groups for safety perceptions, although the age group representation in said work was less granular: below or above 50 years old.
Another work also did not find significant differences in perception scores by age or gender~\cite{Salesses.2013}, although similarly to \cite{Kang.2023}, age groups were analyzed as below or above a given age, i.e., 28 years old in said study.
Unlike work in \cite{Gu.2025}, we did not find significant differences in the \textit{boring} indicator for gender, or any, demographic groups.
Nevertheless, our findings show that in other indicators, both traditional and new, there are significant differences for all five demographic characteristics in different countries.
More importantly, these significant differences are found in cities from continents often overlooked by the research community, e.g., Africa~\cite{Juhász.2024e9f}.
Furthermore, although nested demographic groups presented limitations in sample size, we found significant differences with reasonable sample sizes ($n \geq 20$) in groups that included AHI (Extended data Fig. 1).

Our findings on personality traits follow this same trend (\autoref{fig:personalities-stats}).
With a similar amount of differences compared to those observed across single demographic groups, Abuja (Nigeria), Singapore, and San Francisco and Santa Clara (USA) exhibit differences among residents with distinct personality traits, e.g., personality scores in either Q1 or Q3 of their cities.
Personality traits have been linked with subjective well-being, for example, conscientiousness was found to be a good predictor of life satisfaction~\cite{Hayes.2003}.
Personality groups based on this, and the extraversion trait, led to the most significant differences including when all participants' ratings were combined (middle plot in \autoref{fig:personalities-stats}).
These results open the opportunity to quantify how urban streetscapes are perceived by residents of different levels of well-being.
Despite the remaining three personality traits showing only one significant association, our overall findings reinforce ample literature correlating personality and landscape preference and the relationship of personality and aesthetic judgment~\cite{Abello.1986}.
These links are crucial when designing tailored urban solutions as most of the significant differences over demographic and personality profiles were found within countries, 15 differences compared to four differences across all locations.

Our quantitative study expands the understanding of machine learning model magnitude bias on urban visual perception prediction, corroborates the perception differences from different populations and how different cities are perceived, and quantifies the influence of someone's location when rating their cities (\autoref{fig:perception-comparisons}).
Supporting previous ideas of cities' identity and their characteristics being captured by machine learning models~\cite{Jang.2024, Zhang.2024ec9}, we found that the predictions of ViT-PP2, a model trained on PP2~\cite{Dubey.2016}, differ consistently with our diverse dataset.
While we noted differences between how participants from each city rated all SVI and between how all participants rated each city (top and middle plot in \autoref{fig:perception-comparisons}), the predicted perception scores paint optimistic and pessimistic scenarios for the different cities.
In both scenarios, albeit with some exceptions, the predicted perceptions of a one-size-fits-all and off-the-shelf model are statistically higher than the ground truth scores under positive indicators and statistically lower under negative indicators.
Moreover, participants seem to rely on visual familiarity; influenced by cultural differences, location-specific norms, and preconceived conceptions of the geography; when rating their own city and others.
Despite not having information on the location of the SVI, i.e., the pairwise comparison did not show the city the SVI is from, participants appeared to either be familiar with their city or could identify characteristics similar to their city and use that as a reference when rating other cities they might not be familiar with (bottom plot in \autoref{fig:perception-comparisons}).

We hypothesize that these location-specific visual references provide a perceptual baseline that becomes normalized for participants from specific locations as visual expectations develop differently across urban contexts~\cite{nasar1994urban}.
For example, Dutch residents' expectation of excellent cycling infrastructure may lead them to give lower scores to streetscapes in other cities, where bike facilities rarely match the quality found at home.
Conversely, participants who do not reside in the Netherlands might rely on their preconceptions or expectations of that country, or countries with similar infrastructure, resulting in higher perception scores on imagery from Amsterdam.
For instance, the median SVI perception scores for Amsterdam, rated by all participants, are the highest on all positive indicators across all cities (middle plot in \autoref{fig:perception-comparisons}).
As the majority of participants have lived in their respective cities for more than five years, the unique characteristics of these cities may be ingrained in their inhabitants.
It is also possible that their higher or lower scores could extend to cities culturally or architecturally similar to their own.
For example, participants from Santiago (Chile), might rate SVI from Lima (Peru), with similar scores, given the similarities of both cities.

Our work asserts the importance of multi-city and multi-population analysis.
Previous multi-country work either analyzed only a single country of residency of a dataset~\cite{Salesses.2013}, did not consider participants' location~\cite{Dubey.2016}, or focused on a single feature, e.g., biophilia~\cite{Lefosse.2025}.
Equally importantly, our work evidences that targeted intervention should consider the perception of locals in their own city~\cite{Qiu.2022}, as it will differ greatly if the perceptual evaluation is done using external SVI or by a machine learning model prediction~\cite{Zhou.2025krn}.
Nevertheless, we do not claim generalization or universality of these patterns, and we are limited to evaluating a single city in each country as the scope of this work.
Our contribution lies in thoroughly assessing and quantifying these perceptual differences across multiple locations and populations, highlighting the importance of human-centric and context-aware methodologies for future larger urban visual perceptions studies.
While we handled inter-country biases by analyzing demeaned perception scores in their respective locations (bottom plot in \autoref{fig:perception-comparisons}), in-depth analysis on cross-cultural, location-based biases remains a promising research direction.
At the same time, intra-country differences should be further analyzed, especially in larger and more diverse countries.

Despite the demographic diversity and geographic location of participants, most of our proposed indicators (\textit{live nearby}, \textit{cycle}, and \textit{walk}) capture, predominantly linearly, the trends of the existing six indicators.
These new dimensions are able to expand perception studies with indicators reflecting a user's willingness to be associated with an urban scene (e.g., live, walk, or cycle in the surroundings).
The \textit{green} indicator introduces the opportunity to expand greenery assessment from traditional objective measurements such as pixel-ratio metrics, e.g., Green View Index (GVI), to subjective context-aware human perception data.
While the relationships found with other perceptual indicators were more complex and nonlinear, we found around half of the statistically significant differences in perception scores between demographic and personality profiles (6 out of 13) under these four new indicators.
These highlight the potential of these indicators as input for targeted interventions.
As different demographic groups in diverse location perceive them differently; e.g., different \textit{live nearby} and \textit{cycle} preferences between gender groups in San Francisco and Santa Clara (USA), and Abuja (Nigeria), respectively (\autoref{fig:demographic-stats}); different \textit{green} preferences between high and low agreeableness groups in Singapore; and different \textit{walk} preferences in genders of a given age group in San Francisco and Santa Clara (USA) (Extended data Fig. 1). 
These indicators offer a clearer human subjective perspective that, when complemented with objective measurements, leads to improved urban solutions and assessments~\cite{Qiu.2022, Kang.2023, wang_assess_2024}.
We propose that researchers consider including these indicators where possible and when relevant.

% final conclusion
This work contributes to the few studies in multi-city perception, emphasizes the importance of human-centric data collection, and releases a global and diverse perception dataset.
Methodologically, we introduced a global and, participants-wise, demographically balanced urban visual perception dataset called Street Perception Evaluation Considering Socioeconomics (SPECS) and we thoroughly analyzed perception scores by demographic and personality groups.
Moreover, we quantified and statistically verified the magnitude bias of off-the-shelf machine learning models in perceptual predictions despite being trained on a global dataset like PP2, we isolated and analyzed the effect of how people are influenced by preconceived conceptions of urban contexts, and introduced and validated new indicators relevant to the six commonly used.
Our results draws us closer to understanding in what contexts, e.g., locations and perceptual indicators, urban visual perception is most affected by demographics and personality.
While we are not able to pinpoint a specific number or percentage of how much people's profiles drive perception responses, we found that as such it can lead to significant differences depending on the profiles and locations.
These insights on profiles and location groups can be extended to other urban studies, such as building exterior evaluation for urban development and strategies~\cite{liang_evaluating_2024}. 
Street-level image-based indicators closely correlate with a city's social structure and economic development~\cite{Li.2024g84}.
Thus, as SVI captures the human-level perspective, more closely aligned with people's experiences in cities, studying and understanding urban streetscapes allows researchers and stakeholders to take informed actions~\cite{Lee.2023h3fh}.
While PP2 remains an impressive effort valuable for urban science, our analysis reveals its limitations in localized contexts, suggesting that generalized or one-size-fits-all models may produce distorted results. 
We recommend that researchers fine-tune their perception prediction models with local data, even though this may require additional data collection efforts.
This approach is likely to produce more accurate and better performing models~\cite{Kang.2023, Zhou.2025krn}. 

Although comprehensive and bringing numerous novel results, this study has limitations.
First, online perception surveys, while widely used, present inherent methodological challenges. 
Display resolutions and screen sizes varied among participants, potentially affecting visual perception. 
Even though we standardized our survey interface design across cities and used each country's official language to minimize biases, we could not control which devices participants used. 
Furthermore, online platforms inherently exclude populations with limited technological access or familiarity. 
While our inclusion criteria only specified participants older than 21 years, we did not screen for color blindness or other visual impairments, which may have affected perceptions beyond the \textit{green} indicator. 
Following established practices in perception studies, we did not provide explicit definitions for perceptual indicators, potentially allowing varied interpretations (e.g., ``safe'' from criminals versus from vehicles).

Secondly, we acknowledge the limited number of cities and countries chosen as well as the respective number of images.
Our approach of one city per country per continent limits the generalizability of our findings to the broader urban landscape worldwide.
To mitigate these limitations, we employed stratified sampling of both urban environments (image selection) and participant demographics, ensuring proportional representation of local socioeconomic contexts within each selected city.
Additionally, to ensure statistical power across all ten different indicators for demographic and personality groups' perception ratings analyses, we constrained the number of images per city to 80, to collect enough participant ratings per image and per group to support robust comparisons across multiple demographic and personality dimensions.
Rather than claiming global or country-level representativeness and insights, our study focuses on determining whether perceptual indicators vary systematically across demographic groups in diverse urban environments.
The reduced number of images per city, taken primarily from city centers, might introduce location biases that could affect the interpretation of our findings.
These location biases may include overrepresentation of tourist areas, commercial districts, or architecturally distinctive features that may not reflect typical residential or suburban environments within each city, e.g., very unique urban forms that people associate with specific cities, such as Amsterdam's canals.
While our stratified sampling approach helps ensure demographic representativeness within each city, it cannot fully address the geographical limitations of our urban sampling strategy.
Nevertheless, our contributions rely on thorough comparisons of perceptions across different demographic groups (top and middle plots in \autoref{fig:demographic-stats}), with analyses that control for location when possible (bottom plot in \autoref{fig:perception-comparisons}). 
This methodological approach allows us to isolate demographic effects on urban perception across multiple diverse urban contexts, i.e., five cities in five different continents compared to perception studies routinely using one or two cities~\cite{Zhang.2018j3c, Kang.2021guq, gao_pedaling_2025, wang_assess_2024}.

Finally, we employed a reduced number of pairwise comparisons per image per indicator.
While~\cite{Gu.2025} recommends 22 comparisons for robust Q scores, a threshold we met in the single-city SVI and multi-city participants (middle plot in \autoref{fig:perception-comparisons}), this was not feasible for analyses with reduced participant populations due to demographic filtering. 
We established a minimum threshold of four pairwise comparisons per image per indicator, matching the average reported in the seminal PP2 dataset~\cite{Dubey.2016}. 
Demographic representation was only uniform for gender and age groups, with race and ethnicity being particularly imbalanced, i.e., only 0.5\% of participants self-identified as Native American or Alaska Native, and 7\% selected ``A race/ethnicity not listed here'' across all five cities. 
Given these constraints, we avoid making decisive conclusive statements or claiming global trends.
Instead, we hope our findings encourage researchers to explore previously disregarded paths, particularly incorporating human-related data for spatial and visual perception analytics in countries and continents underrepresented in the literature.

Future research should prioritize analyzing interactions between multiple demographic factors, e.g., older adults with different walk abilities~\cite{Chen.2024ox}, and their combined effects on various perceptual indicators. 
As our study grouped participants by multiple demographics factors, the resulting smaller participant pools yielded fewer ratings per perceptual indicator (e.g., safety perception scores based solely on older women's ratings). 
To address this limitation, subsequent studies could either increase participant numbers or ratings per image. 
The latter could be achieved by reducing the number of images per city, though potentially compromising urban landscape diversity, or by focusing on fewer perceptual indicators, such as exclusively studying safety perception.
Another alternative is to leverage scalable perception rating methods.
Human-machine frameworks~\cite{Yao.2019} allows for rapid and cost-effective perception assessments.
Building on our findings on the demographic and location-specific perceptual differences, these methods could be more appropriately deployed to quickly evaluate entire cities.

Regarding subjective perception modeling, future research could expand on mimicking human perception of an urban scene with Large Language Models (LLMs).
Recent work has begun exploring scene captioning for LLM-based perception of the traditional six indicators~\cite{Ma.2023}, multimodal LLMs for image-to-perception scoring of urban attractiveness~\cite{Malekzadeh.2025}, and landscape evaluation~\cite{Tung.2025}. 
These results show that LLMs need local context understanding to match human evaluative nuances~\cite{Malekzadeh.2025}, reinforcing the importance of understanding specific influence of local factors in our study. 
Our dataset and findings could be leveraged to model demographically and personality specific synthetic participants and inform LLM-based agents, as attempted in~\cite{Verma.2023}, with empirical data.

%%===================================================%%
\section{Methods}\label{sec:methods}
\subsection{Data collection}\label{sec:data-collection}
In our research, we followed a multi-criteria approach that balanced geographic representation, data availability, and cultural diversity while working within practical constraints.
Our selection methodology employed a three-tier filtering process: (1) continental representation to ensure global coverage, (2) availability in established datasets (MIT PP2~\cite{Dubey.2016} and NUS Global Streetscapes~\cite{Hou.2024}) to enable comparative analyses, and (3) consideration of cultural, urban form, and socioeconomic diversity within these constraints. 
This systematic approach reduced our candidate cities considerably (Global Streetscapes includes 688 cities around the world, and PP2 only covers 56) but ensured scientific rigor and comparability with existing literature.
To ensure geographic diversity (1), we included the top five largest continents by population: South America, Europe, Africa, Asia, and North America.
In terms of data availability (2), in South America, only three and two cities from Brazil and Chile are present in PP2, respectively.
We chose Santiago as it was the only capital among the two countries listed in both PP2 and Global Streetscapes datasets, and a better cultural representation of a Hispanic city in South America.
On one hand, while the number of candidate cities in Europe, Asia, and North America is higher, we opted for cities that are commonly used in perception studies due to their unique urban form and cultural characteristics~\cite{Gu.2025, liang_evaluating_2024} (3): Amsterdam, Singapore, and San Francisco.
On the other hand, only a handful of cities from South Africa and Botswana in Africa are present in PP2.
While we aimed to prioritize the overlap between the cities in PP2 and the Global Streetscapes datasets, we did not consider Cape Town and Johannesburg, in South Africa, as representative candidates for the continent.
Unfortunately, the other candidate city, Gaborone in Botswana, presented logistical difficulties when recruiting participants.
Therefore, among the top five countries based on GDP in the continent (South Africa, Egypt, Algeria, Nigeria, and Morocco)~\cite{IMF2024}, we opted for Abuja in Nigeria purely based on its central location in the continent and its presence in the Global Streetscapes.
This city selection, despite its limitations, provides sufficient diversity in urban morphology, cultural contexts, and socioeconomic conditions (Supplementary material Fig. 1) to test our core hypothesis about demographic influences on urban perception. 
We followed the image download process detailed in~\cite{Hou.2024} and downloaded all available street view images within a 2.4 km$^2$ square in the city center.
For the city of Abuja, Nigeria, we manually selected a 2.4 km$^2$ region with more urban development since the images in the Global Streetscapes dataset contained mostly highways scenes.

Based on the contextual metadata available in the dataset, we selected only images with good quality, no reflection, non-panoramic, captured in clear weather, and taken during the day with front or back view direction.
For image validity screening, we used visual complexity, as suggested in the Global Streetscapes dataset~\cite{Hou.2024}.
A low visual complexity value means the image is composed of predominantly one element, e.g., a road or wall.
Supplementary material Fig. 2 shows the distribution of visual complexity values for all filtered images grouped by city.
Images from all cities but Abuja have a clearer overlap in their distribution, with a visual complexity mean of 1.7; thus, we chose a conservative threshold of 1.5 and retained all images with at least this visual complexity value.
However, images from Abuja have lower visual complexity values, with a mean of 1.3.
While this overall lower visual complexity may reflect the intrinsic urban landscape of the selected region in the city, we aimed to have images with similar characteristics across all cities.
Therefore, we retained images with at least a visual complexity of 1.3 as a trade-off for image diversity; there are twice the number of images with a visual complexity $\geq1.3$ than with a visual complexity $\geq1.5$.

From the resulting dataset, and inspired by the work in~\cite{Gu.2025}, we used their segmentation data and clustered them into four clusters that effectively represent images dominated by roads, vegetation, cars, buildings, and sky regions in similar proportions.
We sampled approximately 15-16 images from each cluster within each city, resulting in 80 images from each city.
This stratified sampling was repeated multiple times to make sure the images reflected urban scenes and not highways, deserted roads, etc.
Our final dataset for this study includes a total of 400 images.
While this final number of images may seem relatively low, our city and image selection process ensures diversity across urban landscape features while maintaining statistical power for our primary research question.
Image 1 in Fig.~\ref{fig:methodology-panel} bottom panel illustrates the data collection step for this research framework.

\subsection{Perception survey} 
The survey was reviewed and approved by the Institutional Review Board (IRB) of our university, and the participants were compensated financially.
It was deployed on an online platform and distributed by a third-party local market research vendor.
Using a single data collection vendor with global presence allowed us to recruit 200 participants from each of the five countries mentioned in the above subsection and ensure validity of data and demographic balance, and guarantee consistency in recruiting and data collection.
Considering the number of images, this number of participants was chosen primarily to gather enough demographic groups for downstream analyses of their survey responses.
The complete survey consisted of two different sections; the first section captured the demographics in 11 questions, and the second consisted of 50 unique pairwise comparisons of the 400 images across 10 subjective indicators.
Image 2 in the bottom panel of \autoref{fig:methodology-panel} shows the different demographic moderators and human perception rating indicators. 

\subsubsection{Demographic survey}
Although the cities mentioned in Section~\ref{sec:data-collection} were prioritised, we also considered nearby cities and provinces to gather enough participants.
Specifically, San Francisco and Santa Clara were targeted for participants in the United States of America (USA), the greater Amsterdam area was targeted for participants in the Netherlands, and Nigeria was targeted at a nationwide level.
The demographic questions included gender, age, nationality, country, and city of residence, length of stay in the said city, annual income level, number of household members, and race and ethnicity.
The Big Five Inventory-10 (BFI-10)~\cite{Rammstedt.2007}, a 10-item short version of the Big Five Inventory, was included as the last question in this survey section.
This personality survey has a more colloquial wording of its questions and is sufficient for research settings with limited time constraints~\cite{b5p-gosling}.
Using this questionnaire version helped us minimize participants' survey fatigue, especially since we have asked several other questions beyond those related to personality.
The Big Five personalities have been shown to influence subjective measures in the built environment and lead to clusters of users with different preferences~\cite{Quintana.2023}.

Participants were recruited from the general population, following their national representation by age (adults over 21 years) and gender for each country, with soft quotas for income level.
The latter means that for the greater Amsterdam area, Singapore, and San Francisco and Santa Clara, participants were recruited such that their annual income ranges are close in quantity.
For Chile and Nigeria, participants were recruited on a natural fallout basis.
The available answers for the annual income ranges were adjusted to each country's currency and situation based on the vendor's suggestions, and the entire survey was translated into the official language in each country where it was deployed.
Participants who did not reside in either of the five countries were excluded.
A complete list of questions and their possible answers is listed in Supplementary material Section 2.

\subsubsection{Human perception rating} \label{subsubsec:perceptionsurvey}
The set, and subset, of six indicators for urban perception (\textit{safe}, \textit{lively}, \textit{wealthy}, \textit{beautiful}, \textit{boring}, and \textit{depressing}) proposed in the PP2 dataset~\cite{Dubey.2016} have been widely and consistently adopted in related work on urban perception and human preferences~\cite{Cui.2023c2p, Kang.2023, meir_understanding_2020, hidayati_how_2020, ogawa_evaluating_2024, liang_evaluating_2024}. 
These traditional six perceptual indicators mainly capture people's instantaneous and intuitive impressions of a place.
Thus, we used them to facilitate comparative analysis with previous and future studies.
However, we proposed and included four new indicators: \textit{live nearby}, \textit{walk}, \textit{cycle}, and \textit{green}.
These additional indicators of living preference, cycling, walking, and greenery are presented as the following preference questions: ``Which place looks like a place you want to \textit{live nearby}?'', ``Which place looks better to \textit{cycle}?'', ``Which place looks better to \textit{walk}?'', and ``Which place looks \textit{greener}?'', to capture a proxy of belonging, bikeabiltiy, walkability, and subjective greenery, respectively.
In contrast to the six traditional indicators, these four new ones reflect attitudes toward sustainability, livability, and active mobility, which are the central goals in many current urban studies~\cite{Ito.2024jlo, huang_using_2023, li_measuring_2022}.
Different studies tend to rely on these perceptual indicators as heuristics or proxies for walkability~\cite{huang_using_2023, li_measuring_2022} and bikeability~\cite{Ito.2024jlo}.
Moreover, quantifying and assessing greenery remains crucial for planning and development.
The quality of green spaces has been linked to depression in older adults~\cite{Wang.2025hf}; greenness in urban scenes was found to be a key indicator for assessing the urban renewal potential~\cite{He.2023lk6}; and green spaces offer critical insights for planning at the community level~\cite{He.2025di}.
At the street level, most studies rely on objective measurements such as Green View Index (GVI), e.g., methods that asses pixel-based information, to asses the quality and quantity of green spaces.
However, studies have found that GVI consistently underestimates greenery quantity when compared to subjective greenery evaluation~\cite{Torkko.2023} and subjective and visual measurements were found to contribute the most in the quality assessment of greenery~\cite{Wang.2025hf, He.2023lk6}.
Moreover, \cite{ODESANG2016268} found that greenery could be perceived and valued differently by different genders and age groups, suggesting the need to investigate potential demographic differences in the perception of greenery in our study.
Our proposed indicators are meant to provide a more direct way of computing these dimensions that are commonly sought by the research community and practitioners while maintaining human perspective which has been found to enable and support context-aware urban solutions~\cite{Qiu.2022, Kang.2023, wang_assess_2024}.
To support this, we hypothesize that these four indicators show a positive correlation with the positive indicators \textit{safe}, \textit{lively}, \textit{wealthy}, and \textit{beautiful}; and a negative correlation with the negative indicators \textit{boring} and \textit{depressing}.

In the online survey, each participant rated five random unique pairs of images per indicator, totaling 50 pairwise comparisons across all 10 indicators.
We chose the pairwise comparison scoring method because it has been shown to produce more stable results in urban visual perception studies~\cite{Gu.2025}.
Supplementary material Fig. 3 shows a screenshot of the online survey for one pairwise comparison under the \textit{lively} indicator.
Participants clicked on the image they preferred as an answer to the indicator question, e.g., the question ``\textit{Which place looks \textbf{livelier}}'' in the screenshot, or chose ``Both are the same to me'' for an equally rated answer.
Participants were not given a predetermined definition of the perceptual indicators.
Instead, we aimed to captured their broad subjective interpretation of the ten different indicators.
The complete perception survey questions are listed in Supplementary material Section 2.2.

\subsection{Dataset}\label{sec:method:dataset}
Our final dataset consists of images showcasing urban scenes from the cities of Santiago, Amsterdam, Abuja, Singapore, and San Francisco.
A total of 400 images, 80 per city, were collected, and 200 respondents from each city (or neighboring areas) were recruited as raters, adding to 1,000 respondents in total.
After deploying the online survey for two months, each participant answered a total of 61 questions, consisting of 11 demographic questions and 50 pairwise comparisons.
We name this dataset Street Perception Evaluation Considering Socioeconomics (SPECS) and release it openly.
\autoref{tab:reporting} shows a summary of our collected dataset in the reporting format proposed by~\cite{Gu.2025} and Supplementary material Table 1 shows the demographic background of the recruited participants from each country.

\begin{table}
    \caption{Survey reporting parameters following the survey design guidelines proposed by~\cite{Gu.2025}}
    \begin{tabular}{p{2cm} p{5cm} p{8cm}}
    \toprule
    \textbf{Section} & \textbf{Reporting Parameters} & \textbf{Ours} \\ 
    \midrule
    Survey Structure & Design, allocation, consistency evaluation & Our survey consisted of a demographic section of 11 questions and 50 pairwise comparison questions. \\ 
    \midrule
    Raters & Demographics, sample size, number of ratings per image, recruitment process & We recruited 1,000 respondents (200 from each country) who reside in the city, or neighboring cities, where the images were extracted -- Santiago, greater Amsterdam area, Nigeria, Singapore, and San Francisco and Santa Clara -- and are above 21 years old.
    An equal distribution of respondents, based on annual income range, was prioritised for all but Chile and Nigeria.
    The gender ratio is 1:1 within countries, and the age group ratio is roughly 1:1:1:2 for 21-29, 30-39, 40-49, and above 50.
    More than 78\% of participants from each country have lived there for more than five years, and less than 5\% have lived there for less than a year. 
    Background information from all participants is shown in Supplementary material Table 1.
    Each participant completed 50 ratings, with each image receiving an average of 25 ratings (min. 10, max. 46) per indicator.\\ 
    \midrule
    Scoring & Scoring methodology, rationale for method selection, evaluative indicators & 
    We selected pairwise comparison as the ranking scoring method due to its proven stability over and fairly quick assessment.
    On the online survey, participants were asked to choose their preference between two images given an indicator.
    To avoid a forced choice scenario that could lead to bias~\cite{forced-choice}, and to be consistent with previous and future studies, the option for an equally rated preference (i.e., ``Both are the same to me'') was shown too.
    We asked participants to rate a random pair of images five times for each of the 10 indicators, making up the 50 ratings per participant.
    Supplementary material Fig. 3 shows a screenshot of the online survey.\\ 
    \midrule
    Image & Image type, image source, rationale for image choice, dataset composition & We chose perspective images with good quality, no reflection, non-panoramic, captured in clear weather, and taken during the day with front or back view direction.
    These types of images have proven rating stability~\cite{Gu.2025} and are easily obtained from open data sources.
    Using the Global Streetscapes, a large open, labelled, processed, and worldwide street-level imagery dataset~\cite{Hou.2024}; we obtained 400 images, 80 from each city -- Santiago, Amsterdam, Abuja, Singapore, and San Francisco, for subjective ratings in the survey.\\ 
    \bottomrule
    \end{tabular}
    \label{tab:reporting}
\end{table}

\subsection{Data analysis}
We calculated the perception scores using the Strength of Schedule (SOS) method, first used in urban perception in the Place Pulse 1.0 dataset~\cite{Salesses.2013}.
This method's score is referred to as the Q score, and it reflects the magnitude of the perception, bounded in $[0, 10]$.
Another commonly used method is the TrueSkill score~\cite{herbrich_trueskill_2006}, but unlike Q scores, it requires a higher minimum number of pairwise comparisons; i.e., 22 and 29, respectively; to reach stable scores~\cite{Gu.2025}.
Details about the Q score calculation are found in Supplementary material Section 4.
Relative scores such as Q scores and TrueSkill scores inherently depend on the ratings considered for their calculations. 
Thus, rather than being an absolute measure of an image's perceptual dimensions, these scores are only comparative; they are only valid within the rating pool in which they were calculated.
As shown in the bottom panel of \autoref{fig:methodology-panel}, step 3, we also included the predicted perception scores for all images using an existing perception model.
This perception model, a Vision Transformer (ViT) developed by~\cite{Ouyang.2023} and used in the Global Streetscapes dataset~\cite{Hou.2024}, serves as a proxy of a global or one-size-fits-all approach as it was fine-tuned with the PP2 dataset~\cite{Dubey.2016}.
Specifically, the model was trained on the top and bottom 5\% rated images for each perception dimension, which were selected as the representative positive and negative samples, respectively, adapting the method by \cite{Zhang.2018j3c}. 
The model was then trained, with a 80\%-to-20\% train-test split, to predict the probability of an image being assigned to the positive label, which was subsequently multiplied by 10 to obtain the perception score.
This model, which we refer to as ViT-PP2, outputs perception scores for the six traditional indicators of the PP2 dataset and reported a higher accuracy in all but the \textit{boring} indicator compared to the model proposed with the PP2 dataset.

We divided the analysis into five parts, as noted in the bottom panel of \autoref{fig:methodology-panel} step 4.
First, we looked at demographic factors and personality traits as moderators in urban visual perception.
For each country, we created groups based on the available answers for each demographic question: gender, age group, annual household income (AHI), education level, and race and ethnicity.
While we tried to keep the original groups on each demographic question, the granularity in some answers presented challenges.
The available groups in AHI are as few as five groups and as many as 12 groups and are different in each country, which would also make it difficult to compare answers between countries.
Thus, we decided to remap existing answers into \textit{Lower}, \textit{Middle}, and \textit{Upper} groups (\autoref{tab:remapping}).
We also remapped some of the answer from the education level question into a common single one.
Answers such as ``Masters degree'' and ``Doctorate or professional degree'' were remapped into ``Postgraduate degree''.

With this variable remapping, we also explored demographics interactions where possible.
Using all five demographic factors (i.e., gender, age, income, education, race and ethnicity) to create nested groups greatly reduces the number of participants within each group.
This reduction limits the available pairwise comparisons to a maximum of one pairwise comparison per image across all indicators and locations.
Therefore, we used only the combinations of `gender $\times$ age group $\times$ AHI' as nested groups.
These demographic factors are commonly explored in urban perception studies~\cite{hidayati_how_2020, Cui.2023c2p, Dubey.2025, Luo.2024} and among our demographic factors, they have the fewest categories; two, four, and three, respectively; resulting in fewer nested groups.
Nested groups based on these three demographic factors yield an average of six pairwise comparisons per image across all indicators and locations, exceeding our selected threshold of four pairwise comparisons used throughout our analyses.

When analyzing each demographic factor in isolation and in interactions, i.e., computing Q scores based on a demographic group, we did not consider ratings from participants who answered ``Prefer not to answer''(39 out of 1,000 participants in AHI), ``Less than Secondary/High school'' and ``Other'' (11 out of 1,000 participants in education level), and ``A race/ethnicity not listed here'' (69 out of 1,000 participants in the race and ethnicity question) due to their small sample size.
On average, these excluded responses would have contributed only one additional rating to the Q score computation of each perceptual indicator for between 5 and 76 images per location.
All subsequent analyses used all participants' ratings.

Personality trait scores were calculated using their respective formula~\cite{Rammstedt.2007}.
To better understand the effect of the personality trait on perception scores, we grouped participants into two groups based on their personality trait score: within the 25$^{th}$ ($\leq$Q1) and within the 75$^{th}$ ($\geq$Q3) percentile, for each personality trait.
This grouping focuses on participants who exhibit the most distinct values of these traits, thus providing a clearer contrast and enhancing the interpretability of their influence on perception scores.

We created these demographic and personality groups for each country, and a set of groups by combining all countries too.
While this nested grouping allowed for more detailed analysis, by reducing the number of available participants in each group, the number of pairwise comparisons per image per indicator was also reduced.
Thus, after computing the Q scores in each nested group, we retained all images with at least four pairwise comparisons, as this was the reported average comparison in PP2~\cite{Dubey.2016}.
Since each demographic group has a different number of samples, i.e., rated images with at least four pairwise comparison by participants with the same demographic characteristics, we prioritized a higher minimum number of samples, i.e., n $\geq 20$, where applicable but reported the minimum used in each demographic group.

To quantify the influence of demographic and personality attributes, we performed statistical tests on the Q scores for all SVIs rated by participants in their respective groups.
Given the difference in sample sizes, i.e., the number of rated SVI by available participants in each group, and normal distribution but unequal variance of Q scores, we chose Welch's ANOVA as the statistical method to compare whether the groups' perception scores means are significantly different.
For demographic attributes with more than two groups, we performed the Games-Howell post-hoc test to identify the significant differences, if any, between each pair of groups' means.

\begin{table}
    \centering
    \begin{tabular}{*{6}{c}}
        AHI bracket & Chile & Netherlands & Nigeria & Singapore & USA\\
        \toprule
        Lower &  \makecell{20,400,000 CLP\\ \& below} & \makecell{29,999 euros\\ \& below} & \makecell{2,000,000 NGN\\ \& below} & \makecell{SG\$0 -\\ SG\$44,999} & \makecell{US\$0 -\\ US\$49,999} \\
        \midrule
        Middle & \makecell{20,400,001 CLP -\\60,400,000 CLP} & \makecell{30,000 euros -\\81,999 euros} & \makecell{2,000,001 NGN-\\4,000,000 NGN} & \makecell{SG\$45,000 -\\ SG\$99,999} & \makecell{US\$50,000 -\\ US\$99,999} \\
        \midrule
        Upper & \makecell{60,400,001 CLP\\ \& above} & \makecell{82,000 euros\\ \& above} & \makecell{4,000,001 NGN\\ \& above} & \makecell{SG\$100,000 -\\ SG\$999,999} & \makecell{US\$100,000 -\\ US\$999,999}\\
        \bottomrule
    \end{tabular}
    \label{tab:ahi-remapping}
    \caption{
    \textbf{Demographic answers remapped values for Annual Household Income (AHI) to handle granular responses.}
    All available answers are listed in the Supplementary material Section 2.
    }
    \label{tab:remapping}
\end{table}

Second, we looked at how people living in the same place perceive different places.
We compared the perception Q scores of imagery from multiple cities rated by participants from a single location, i.e., country.
We also compared the predicted perception scores of all imagery by the ViT-PP2 perception model against the combined ratings from all participants without grouping them by location.
We used the same statistical method from the previous analysis, Welch's ANOVA, to find statistically significant differences, and we also used the same threshold of four minimum pairwise comparisons per image per indicator, and it resulted in at least 71\% of the dataset, per country, usable.
Third, in a similar vein, we compared how people from all five different countries perceived the same place.
As this allowed us to use all participants' ratings, we used a higher threshold of 22 pairwise comparisons.
We used more than 51 images per city (63.75\% of the available imagery per city).
These perception Q scores were also compared with the ViT-PP2 predicted scores for imagery from each city.
We performed these score comparisons with the output of a trained model as a way of comparing our collected ground truth against a one-size-fits-all prediction model.
These comparisons were also based on Welch's ANOVA statistical significance tests.
Fourth, expanding on the second comparison of multi-city imagery and single-city participants, we looked into how people living in the same (or neighboring) city perceive other cities relative to their own.
As this approach is on a similar vein to the second one, we used the same threshold and dataset size.
We analysed the average change in perception of participants rating imagery from their own city compared to imagery from the other four cities.
To debias the responses from cultural and location-based influences, we scaled the perception Q scores to z-scores among ratings for each city and indicator from all participants.
This transformation avoids the cross-cultural perception bias as it looks at the scores within locations.
We only included images with at least four pairwise comparisons per indicator, resulting in more than 50 images with Q scores per city per indicator (62.5\% of the available SVI per city).
Finally, we performed linear, i.e., Pearson correlation coefficient (R), and non-linear, i.e., quadratic (normal and inverted U-shaped) and cubic, correlation analyses between our proposed new four indicators and the existing six traditional ones to validate their usability and interpretability.
For this, we calculated the perception Q scores using the ratings from all 1,000 participants and retained the images with at least 22 pairwise comparisons per indicator for more stable results as suggested by~\cite{Gu.2025}.

%%===================================================%%
\section*{Data availability}
Annotated and labeled SVI data are obtained from the NUS Global Streetscapes dataset (\texttt{https://huggingface.co/datasets/NUS-UAL/global-streetscapes}).
Raw images were obtained from the crowdsourced platforms Mapillary and KartaView.
We share openly our dataset SPECS (Street Perception Evaluation Considering Socioeconomics), consisting of survey responses and participants' demographic data at
\texttt{https://huggingface.co/datasets/matiasqr/specs}.

%%===================================================%%
\section*{Code availability}
The step-by-step process and code for all analyses are available in the public repository: 
\texttt{https://github.com/matqr/specs}.

%%===================================================%%
\section*{Acknowledgements}
We thank the survey participants from around the world for their time and responses.
The authors also thank Annette Gloria Fernandez, Felix Hammer, and Stella Morgenstern for the initial discussions and Milieu Insight Pte Ltd for the data collection work.
This research was conducted at the Future Cities Lab Global at Singapore-ETH Centre. 
Future Cities Lab Global is supported and funded by the National Research Foundation, Prime Minister's Office, Singapore under its Campus for Research Excellence and Technological Enterprise (CREATE) programme and ETH Z\"urich (ETHZ), with additional contributions from the National University of Singapore (NUS), Nanyang Technological University (NTU), Singapore and the Singapore University of Technology and Design (SUTD).
This research was supported by the Singapore International Graduate Award (SINGA) scholarship provided by the Agency for Science, Technology, and Research (A*STAR), the NUS Graduate Research Scholarship, and the NUS.
This research is part of the project Multi-scale Digital Twins for the Urban Environment: From Heartbeats to Cities, which is supported by the Singapore Ministry of Education Academic Research Fund Tier 1.
This research is part of the project Large-scale 3D Geospatial Data for Urban Analytics, which is supported by the National University of Singapore under the Start Up Grant R-295-000-171-133.

%%===================================================%%
\section*{Author contributions}
M.Q.: Conceptualization, Methodology, Software, Validation, Formal analysis, Investigation, Data Curation, Writing - Original Draft, Visualization, Project administration.
Y.G.: Methodology, Investigation, Software, Writing - Review \& Editing.
X.L.: Methodology, Visualization, Writing - Review \& Editing.
Y.H.: Methodology, Data Curation, Writing - Review \& Editing.
K.I.: Methodology, Software, Writing - Review \& Editing.
Y.Z.: Writing - Review \& Editing.
M.A.: Writing - Review \& Editing.
F.B.: Conceptualization, Methodology, Resources, Writing - Review \& Editing, Supervision, Funding acquisition.

%%===================================================%%

\bibliography{demographics-perception}

%%===================================================%%

\end{document}